%% file: main.tex
\documentclass[10pt,twocolumn,letterpaper]{article}
\usepackage[pagenumbers]{cvpr}      
\input{preamble}

\definecolor{cvprblue}{rgb}{0.21,0.49,0.74}
\usepackage[pagebackref,breaklinks,colorlinks,citecolor=cvprblue]{hyperref}
\usepackage{microtype}
\usepackage[section]{placeins}
\usepackage{multirow}
\usepackage{booktabs}
\usepackage{algorithm}
\usepackage{algorithmic}
\usepackage{bm}
\def\paperID{1667} 
\def\confName{CVPR}
\def\confYear{2024}

\newcommand{\x}{{\bm{X}}}
\newcommand{\y}{{\bm{Y}}}

\newcommand{\B}{\textcolor[rgb]{0.00,0.00,0.00}}

\title{High-Resolution Cloud Detection Network}

\author{Jingsheng Li$^1$, Tianxiang Xue$^1$, Jiayi Zhao$^1$, Jingmin Ge$^2$, Yufang Min$^{3,4}$, Wei Su$^1$, and Kun Zhan$^{1,\star}$\\
1. School of Information Science and Engineering, Lanzhou University\\
2. College of Atmospheric Sciences, Lanzhou University\\
3. Northwest Institute of Eco-Environment and Resources, Chinese Academy of Sciences\\
4. National Cryosphere Desert Data Center, Lanzhou,  Gansu, China\\
{\small \url{https://github.com/kunzhan/HR-cloud-Net}}}

\begin{document}
\maketitle

\begin{abstract}
The complexity of clouds, particularly in terms of texture detail at high resolutions, has not been well explored by most existing cloud detection networks. This paper introduces the High-Resolution Cloud Detection Network (HR-cloud-Net), which utilizes a hierarchical high-resolution integration approach. HR-cloud-Net integrates a high-resolution representation module, layer-wise cascaded feature fusion module, and multi-resolution pyramid pooling module to effectively capture complex cloud features. This architecture preserves detailed cloud texture information while facilitating feature exchange across different resolutions, thereby enhancing overall performance in cloud detection. Additionally, a novel approach is introduced wherein a student view, trained on noisy augmented images, is supervised by a teacher view processing normal images. This setup enables the student to learn from cleaner supervisions provided by the teacher, leading to improved performance. Extensive evaluations on three optical satellite image cloud detection datasets validate the superior performance of HR-cloud-Net compared to existing methods.
\end{abstract}

\section{Introduction}
The International Satellite Cloud Climatology Project~\cite{zhang2004calculation} estimates that clouds cover approximately 66\% of the Earth’s surface\cite{king2013spatial}. Cloud detection is crucial for remote sensing, as clouds frequently obscure surface details, thereby degrading the quality and utility of satellite imagery. Thus, accurate cloud detection is essential to maintain the integrity of satellite imagery.

The growing accessibility of optical satellite data has led to the emergence of numerous cloud detection algorithms~\cite{zhu2012object,zhu2015improvement}. The cloud detection task generally refers to the binary segmentation of clouds in remote sensing images. U-Net~\cite{ronneberger2015u}, known for its effectiveness in binary segmentation tasks, provides a strong foundation for cloud detection. However, \textit{vanilla} U-Net has limitations, particularly in the fusion of feature maps between the encoder and the decoder, restricted to the same resolution. This constraint poses challenges in effectively leveraging various resolutions for intricate target shapes, edges, and textures, impeding the ability of \textit{vanilla} U-Net to utilize high-resolution information crucial for accurate object detection.

To address the limitations of using \textit{vanilla} U-Net for cloud detection and to tailor it for this specific task, several advanced methods have been developed. CDNet~\cite{yang2019cdnet} employs dilated convolutions in the encoding stage and introduces a feature pyramid module and edge-refinement modules to enhance U-Net. CDNet-v2~\cite{9094671} further enhances U-Net, introducing skip connections in the decoder to improve the integration of finer details from lower-level features. This adaptation effectively addresses challenges related to the variable size and irregular structure of clouds, optimizing cloud detection capabilities. U-Net3$+$~\cite{huang2020Unet}, as applied in cloud detection~\cite{Yin2022}, employs nested and dense skip connections to fuse feature maps of different resolutions. CDU-Net~\cite{rs13224533} introduces a high-frequency feature extraction network to capture shallow multi-resolution information to refine cloud boundaries and predict fragmented clouds. GANet~\cite{du2023gated} adopts a fully connected interaction structure to leverage both high-dimensional and low-dimensional features, filtering out redundant information. However, these methods do not sufficiently focus on incorporating the shallow high-resolution information necessary for a more detailed description of cloud textures.

Distinct from semantic segmentation tasks on natural images, cloud texture details vary significantly within each type due to their inherent characteristics and interaction with other environmental factors. Consequently, clouds exhibit distinctive texture and structural features. Neglecting attention to these texture features may compromise the accuracy of cloud detection. High-resolution images comprehensively capture texture details crucial for forming cloud patterns and structures. Through a thorough exploration of high-resolution information, we aim to precisely describe the micro-structure and texture details, enhancing the accuracy of cloud detection and understanding. Building on these observations, we are inspired to design a network capable of exploring and integrating high resolutions in cloud images. This design is specifically crafted to effectively tackle the distinctive challenges associated with cloud detection. Inspired by the success of high-resolution networks~\cite{hrnet8953615,wang2020deep} and U-Net~\cite{ronneberger2015u}, particularly for tasks requiring detailed information, we propose a novel neural network architecture named High-Resolution Cloud Detection Network (HR-cloud-Net). This architecture integrates features such as parallel multi-resolution representation encoding, layer-wise cascaded feature fusion, and utilization of multi-resolution pyramid pooling for integrating high-resolution information. Additionally, we implement an effective multiview training strategy, introducing a student view trained on augmented images while utilizing the teacher view trained on normal images to supervise the student view. This provides a robust solution for cloud detection tasks and highlights the potential for achieving effective generalization with limited data. To evaluate performance of HR-cloud-Net, experiments are conducted on the CHLandSat-8 dataset~\cite{du2023gated}, the 38-cloud dataset~\cite{8898776}, and the SPARCS dataset~\cite{rs6064907}. HR-cloud-Net achieves superior performance across various evaluation metrics.

The main contributions are summarized as follows:
\begin{itemize}
	\item We introduce HR-cloud-Net, which integrates a high-resolution representation module, a layer-wise cascaded feature fusion module, and a multi-resolution pyramid pooling module.  The objective is to effectively mitigate semantic gaps in the feature extraction process across different resolutions of cloud textures.
	\item We introduce a novel multiview training strategy. The input image is split into two views, with high-confidence predictions from the teacher view supervising the student view. The teacher view is trained on normal images, while the student view is trained on augmented images. This carefully designed multi-view training process enhances generalization performance of HR-cloud-Net.
	\item In the three datasets, HR-cloud-Net achieves the comparable performance of other state-of-the-art models. Based on experimental results, we find that HR-cloud-Net is effective in cloud detection, and our multiview training strategy enhances its performance.
\end{itemize}
\section{\B{Related Work}}
\B{Cloud detection methodologies can be broadly categorized into three types: Classic methods, U-Net-based methods, and deep learning algorithms that explore multiple resolutions.}
\subsection{\B{Classic Cloud Detection Algorithms}}
\B{The field of cloud detection has evolved from traditional threshold-based approaches to more sophisticated methods. Initially, algorithms like Fmask~\cite{zhu2015improvement} utilized decision trees based on distinctive cloud features in various spectral bands. However, these methods were limited by their sensitivity to lighting conditions and specific data characteristics.}

\B{To improve detection accuracy, methods observing different time periods were developed. For instance, CCDC~\cite{zhu2014continuous} leverages multispectral satellite imagery from multiple time points, using temporal analysis for effective cloud detection. Despite its effectiveness, this method demands high data storage and processing capabilities and is susceptible to cloud cover and seasonal variations, adding complexity and potential errors.}

\B{The limitations of traditional methods have been addressed by advancements in deep learning for remote sensing~\cite{li2018cloud}. Deep neural networks enable end-to-end learning and feature extraction, resulting in efficient and accurate cloud detection. These technological advancements continue to improve model performance~\cite{rs11222719}.}
\subsection{\B{U-Net Cloud Detection Algorithms}}
\B{Recent years have seen substantial progress in cloud detection through deep learning in remote sensing~\cite{li2018cloud,rs11222719}. Among the various deep learning models, U-Net~\cite{ronneberger2015u} and its variants have shown outstanding performance in image segmentation tasks, especially in scenarios with limited training samples and the need for detailed information recovery. Originally designed for medical image segmentation, U-Net's symmetric encoder-decoder structure with skip connections facilitates improved detail recovery and has been effectively adapted for cloud detection.}

\B{RS-Net~\cite{jeppesen2019cloud} applies the U-Net architecture to achieve high-precision cloud detection. CDNet~\cite{yang2019cdnet} improves upon U-Net by incorporating a deeper ResNet-50~\cite{he2016deep} model with feature pyramids and edge-refinement modules, achieving high-precision cloud detection in satellite thumbnails. CDNet-v2~\cite{9094671} further enhances this by introducing adaptive feature fusion and advanced semantic information guidance modules. U-Net++~\cite{zhou2018Unet++}, another variant, emphasizes multi-level feature integration through nested U-Net sub-networks, highlighting the importance of multi-resolution features.}
\subsection{\B{Cloud Detection with Multiple Resolutions}}
\B{Achieving precise cloud segmentation relies on the effective integration of multi-resolution features. While previous networks have demonstrated some effectiveness, they often fall short in capturing comprehensive multi-resolution information.}

\B{To address this challenge, various techniques are explored to enhance the exploitation of multi-resolution features. U-Net3+~\cite{huang2020Unet}, widely employed in cloud detection~\cite{Yin2022}, leverages full-scale skip connections to amalgamate high-level semantic information from feature maps of varying scales, thereby achieving more precise segmentation. CDU-Net~\cite{rs13224533} introduces a high-frequency feature extraction network to capture shallow, high-resolution information, complemented by multi-resolution convolutions for refining cloud boundaries and predicting fragmented clouds. PSPNet~\cite{zhao2017pyramid} captures semantic information across different resolutions, enabling precise pixel-level segmentation. HRNet~\cite{hrnet8953615,wang2020deep}, addressing challenges in high-resolution image tasks, demonstrates exceptional performance in image segmentation and key-point detection through its high-resolution feature pyramid and multi-resolution fusion mechanism. CRSNet~\cite{zhang2023crsnet} designs a hierarchical multi-scale convolution module to explore multiple resolutions. MCANet~\cite{hu2023mcanet} employs a multi-branch network for segmenting cloud and snow regions in high-resolution remote sensing images. MAFANet~\cite{10144804} designs a multi-scale attention feature aggregation and a multi-scale stripe pooling attention module to exploit multi-resolution representation.}

\B{This paper emphasizes the importance of simultaneously preserving high-resolution feature recovery and applying a skip-structure upsampling process to low-resolution feature maps. The integration of a multi-resolution pyramid pooling mechanism recaptures diverse resolution semantics, resulting in superior performance compared to baselines. Our proposed method aims to address the limitations of previous networks, particularly in achieving more accurate segmentation in complex cloud detection scenarios.}
\section{HR-cloud-Net for Cloud Detection}
This section outlines the architecture of HR-cloud-Net and introduces a novel multiview supervision strategy employed by HR-cloud-Net. We propose a more comprehensive and accurate cloud detection approach, consisting of two main components:
(1) A well-designed neural network, HR-cloud-Net, adept at accurately identifying cloud regions, and (2) a multiview supervision strategy aimed at improving generalization performance.

The dataset used in this article comprises a collection of large remote sensing images. These large images are uniformly partitioned into smaller segments of size $H \times W$, resulting in the creation of a training dataset $\mathcal{D} = \{\x_1, \ldots, \x_n\}$. Each small image, denoted as $\x \in \mathbb{R}^{H \times W\times3}$ with three channels and a size of $H \times W$, undergoes augmentation to create a corresponding augmented image $\x^{\text{aug}}$. Both $\x$ and $\x^{\text{aug}}$ are then simultaneously input into HR-cloud-Net $f(\cdot | \theta)$, where $\theta$ represents the HR-cloud-Net parameters. HR-cloud-Net produces outputs $\y = f(\x | \theta)$ and $\y^{\text{aug}} = f(\x^{\text{aug}} | \theta)$, which are utilized in our multiview supervision strategy. The prediction $\y\in\mathbb R^{H\times W\times2}$ has two channels, the first one predicts the background and the second predicts the cloud. The coresponndiing ground-truth label is dented by $\bm T\in\mathbb R^{H\times W\times2}$.
\subsection{HR-cloud-Net}
\begin{figure*}[!t]
	\centering %
	\includegraphics[width=0.96\textwidth]{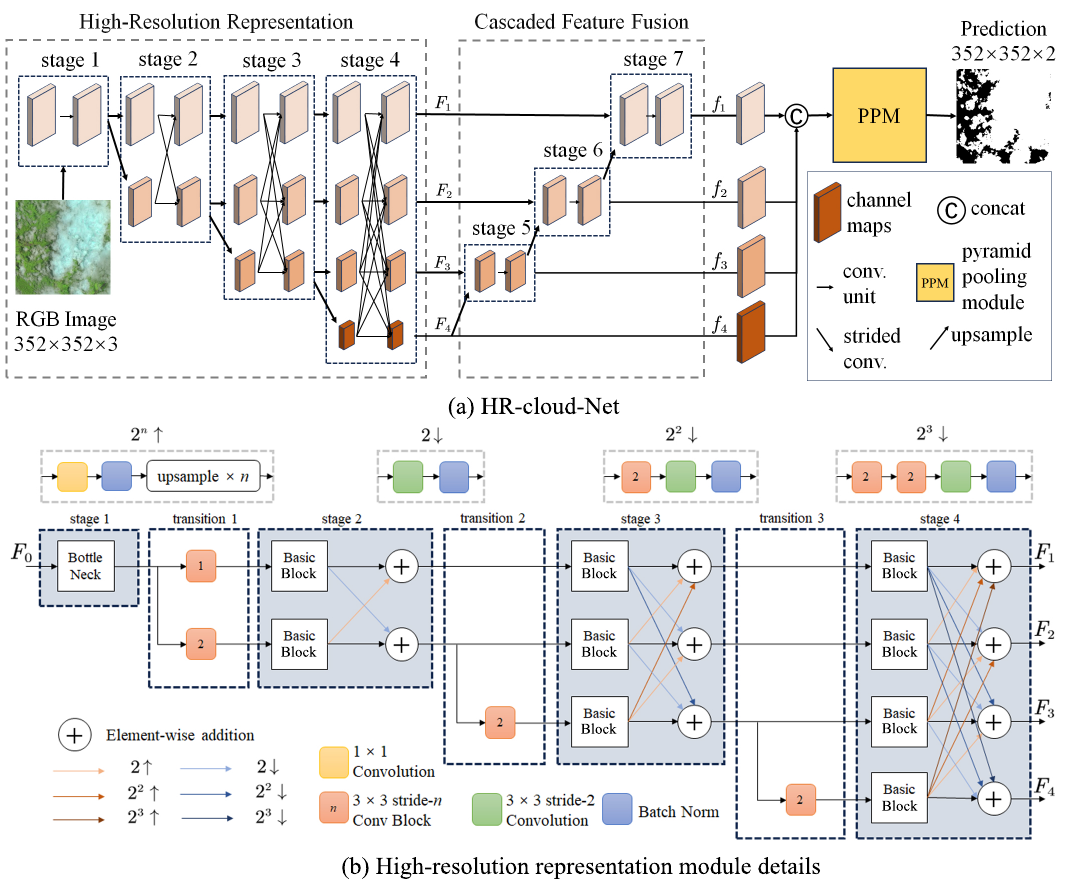}
	\caption{High-resolution cloud detection network, HR-cloud-Net. (a) is the overall architecture of HR-cloud-Net, and (b) is high-resolution representation module details. Bottle Neck and Basic Block are derived from ResNet\cite{he2016deep}.}
	\label{architecture}
\end{figure*}

For cloud detection within remote sensing images, HR-cloud-Net is tasked with several objectives: efficiently locating cloud pixels, distinguishing their irregular shapes from similar background elements such as snow and lakes, and precisely delineating intricate cloud boundaries, including small and thin cloud regions. To address these challenges, we propose HR-cloud-Net, a cloud detection network designed to accurately identify cloud regions.

As shown in Fig.~\ref{architecture}, HR-cloud-Net adopts an asymmetric encoder-decoder structure. Initially, multi-level features are extracted from remote sensing images, yielding image features $F_1, F_2, F_3$, and $F_4$ at various resolutions~\cite{hrnet8953615,wang2020deep}. Subsequently, these features undergo layer-wise layer-wise cascaded feature fusion~\cite{ronneberger2015u}. The resulting output features $f_1, f_2, f_3$, and $f_4$ are upsampled to the same size and concatenated along the channel dimension. To further enhance the concatenated image features and extract global contextual features, a multi-resolution pyramid pooling module~\cite{zhao2017pyramid} is employed. Finally, the processed feature maps undergo channel compression convolution to produce the final predicted cloud area of HR-cloud-Net.

The high-resolution representation module of HR-cloud-Net comprises four stages, each complemented by transition modules. These stages maintain high-resolution representations by iteratively integrating representations from multiple resolutions, ensuring the generation of reliable high-resolution outputs. Meanwhile, the transition module facilitates the creation of new resolution features. The image $\x$ initially undergoes two convolutional layers with a $3\times3$ kernel size and a stride of 2, resulting in feature map $F_0$ with a resolution of $\frac{1}{4}$ of $\x$. $F_0$ serves as the input of stage 1. With the top four stages, HR-cloud-Net generates encoded representations $F_1, F_2, F_3$, and $F_4$, corresponding to four resolution levels. $F_1$ has the resolution consistent with $F_0$, while $F_2, F_3$, and $F_4$ have resolutions of $\frac{1}{8}$, $\frac{1}{16}$, and $\frac{1}{32}$ of $\x$, respectively.

The first stage of HR-cloud-Net includes a single high-resolution feature, gradually incorporating low-resolution features into subsequent stages for parallel feature extraction and fusion. Consequently, the feature resolution in the subsequent stages include one more lower resolution compared to the previous stage. The detailed high-resolution representation module, shown in Fig.~\ref{architecture}(b), comprises four parallel features. Each Basic Block in parallel feature consists of four residual units arranged in parallel, derived from ResNet~\cite{he2016deep}. The outputs of the four residual units are finally fused within the Basic Blocks. By adopting this structure of Basic Blocks, information loss during information exchange between different resolutions can be reduced.

The transition module selects the feature map with the lowest resolution among the outputs of the stage after stages 1, 2, and 3, respectively. It then applies downsampling convolutions with a stride of 2 to generate new resolution features, which are subsequently combined with other parallel resolution features as input for the next stage.

To illustrate the process of the multi-resolution representation module, let's consider stage 2, shown in Fig.~\ref{architecture}(b). At this stage, the inputs undergo ``Basic block'' operations,  resulting in the generation of $H_1$ and $H_2$. Subsequently, the corresponding outputs, denoted as $G_1$ and $G_2$, are produced. Each output is the sum of the transformed inputs from two resolution representations. Specifically, $G_1$ is calculated by $G_1={\rm ReLU} \bigl(H_1+T_{21}(H_2)\bigr)$. Here, $T_{21}(H_2)$ signifies a series of operations involving a $1 \times 1$ convolution layer, followed by batch normalization, and bilinear upsampling to match the size of $H_1$. Then, $G_2$ is derived from $H_2$ and $H_1$: $G_2={\rm ReLU} \bigl (H_2+T_{12}(H_1)\bigr)$. Here,  $T_{12}(H_1)$ involves downsampling $H_1$ using a stride-2 $3 \times 3$ convolutional layer, followed by batch normalization. 

The high-resolution representation module comprises four stages, each fully leveraging detailed features at different resolutions during the transition from low resolution to high resolution. We adopt a parallel connection approach for the high-to-low resolution features in the encoding stage, diverging from a serial connection~\cite{hrnet8953615,wang2020deep}. By iteratively performing multi-resolution fusion~\cite{hrnet8953615,wang2020deep}, HR-cloud-Net facilitates information exchange among feature maps in different resolutions,  promoting mutual enhancement between low-resolution and high-resolution representations. This iterative process yields a more robust semantic representation of clouds across different resolutions, thereby enabling accurate delineation of general cloud areas. Importantly, this architectural decision ensures that HR-cloud-Net maintains high resolution throughout the feature extraction process, allowing it to capture distinctive cloud textures and structural features with greater spatial precision and richer detail. As a result, HR-cloud-Net exhibits an improved ability to delineate cloud boundaries while avoiding overlooking thin or small clouds.

$F_1, F_2, F_3$, and $F_4$ obtained from the high-resolution representation module undergo three key steps to generate the final predicted binary cloud mask: (1) layer-wise cascaded feature fusion, (2) parallel multi-branch feature concatenation, and (3) multi-resolution pyramid pooling module, as shown in Fig.~\ref{architecture}(a).

The cascaded feature fusion comprises three stages, each involving bilinear interpolation upsampling, feature concatenation, and two convolutional blocks. Taking stage 5 in Fig.~\ref{architecture}(a) as an example, its input includes $F_3$ from the encoding stage at the current resolution and $f_4$ at a lower resolution than $F_3$. For the lowest resolution, $f_4 = F_4$. Upsampling $f_4$ aligns its resolution with $F_3$ while maintaining the same channel number. It is then concatenate with $F_3$ along the channel dimension, resulting in features with the same resolution as $F_3$ and a channel number equal to the sum of channels in $F_3$ and $F_4$. Finally, it undergoes two convolutional blocks denoted as $f_3 = {\rm ConvBlock}\bigl({\rm ConvBlock}\bigl([F_3, f_{4}^{2\uparrow}]\bigr)\bigr)$, where $F_3$ and $f_{4}^{2\uparrow}$ are concatenated along the channel dimension, and ${\rm ConvBlock}$ represents a convolutional block consisting of a $3\times3$ convolution, batch normalization, and ReLU. The resulting aggregated feature $f_3$ serves as input for the upsampling layer of the stage 6 and as the output of the current resolution feature. This process continues with cascaded feature fusion between adjacent scales until the final output results are obtained.

\begin{figure}[htbp]
	\centering %
	\includegraphics[width=0.5\textwidth]{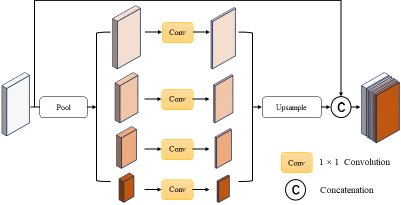}
	\caption{Multi-resolution pyramid pooling module.}
	\label{PPM}
\end{figure}

We upsample the three parallel resolution features $ f_1, f_2 $, and $f_3$ obtained previously to match the resolution of $ f_1 $. Subsequently, we concatenate all the feature layers of the same size along the channel dimension. This fusion process creates a comprehensive feature layer that preserves both shallow and deep positional and semantic information. By retaining edge features of the feature maps, HR-cloud-Net effectively focuses on global features without sacrificing local details.

To enhance HR-cloud-Net's ability to utilize global information and capture global contextual features, we incorporate a multi-resolution pyramid pooling module~\cite{zhao2017pyramid} at the end of HR-cloud-Net. As shown in Fig.~\ref{PPM}, this module processes image features through four parallel pooling layers of varying resolutions. Subsequently, the channel number is reduced to $\frac14$ of the original using $1 \times 1$ convolution layers for each resolution. These compressed features are then concatenated with the original feature map along the channel dimension. Finally, the feature maps produced by the multi-resolution pyramid pooling module are condensed into two channels to yield HR-cloud-Net's final predicted cloud area. After an additional upsampling step, the predicted image is restored to match the size of the original input remote sensing image, as depicted in Fig.~\ref{architecture}(a). In HR-cloud-Net's final prediction, the first channel indicates the probability of being classified as background, while the second channel represents the probability of being classified as cloud.

Due to the presence of background elements such as snow, ice, and lakes in certain remote sensing images, which can be easily confused with clouds, HR-cloud-Net faces challenges in accurately delineating cloud boundaries. This difficulty arises from the limited depth of semantic features provided by high-resolution shallow spatial structure features, making it challenging to differentiate foreground and background information. However, our layer-wise cascaded feature fusion module and multi-resolution pyramid pooling module, addresses this issue by consistently emphasizing the fusion of shallow and deep semantic information.

In the cascaded feature fusion module, the output features at the highest resolution incorporate semantic information from multiple parallel layers, followed by the aggregation of all parallel layer features. This mechanism enables HR-cloud-Net to accurately identify the semantic boundaries of clouds while effectively filtering out background information like snow, ice, and lakes. Furthermore, the multi-resolution pyramid pooling module enhances this process by integrating contextual information from the image, consolidating details across various depths and scales while preserving high-resolution feature information. This approach helps mitigate interference from redundant information, such as cloud textures, thereby enhancing HR-cloud-Net's effectiveness in reconstructing details and improving its ability to detect cloud regions accurately.
\begin{figure}[htbp]
	\centering
	\includegraphics[width=0.5\textwidth]{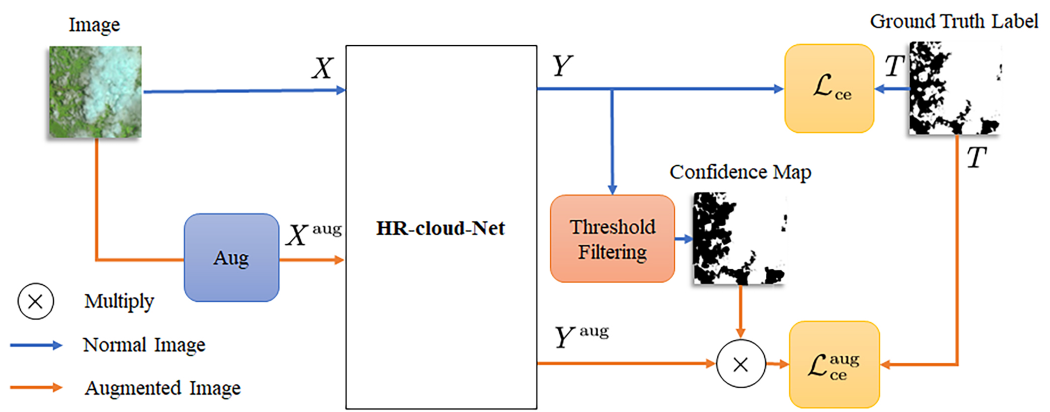}
	\caption{Multiview supervision of HR-cloud-Net. Aug denotes the data augmentation, and CE Loss is the cross-entropy loss function. The blue line represents the view of normal image feature, while the orange line represents the augmented image view.}
	\label{MST}
\end{figure}
\subsection{Multiview Supervision}
\B{As shown Fig.~\ref{MST}, we feed HR-cloud-Net two-view inputs, one is the image itself and the other is a strong augmented view. For each view, we use different loss function. The main difference between the two terms lies in their update mechanisms. The first term uses standard cross-entropy and updates the model with the entire prediction, whereas the second term selects high-confidence predictions. Both are based on cross-entropy, but one applies to the whole prediction while the other employs a selection strategy. High-confidence predictions supervise the strongly augmented view to enhance generalization. Since our approach focuses on discriminating cloud pixels, cross-entropy is used for both terms in the overall loss function. The weak view, which has higher confidence, supervises the strong augmentation view, necessitating a threshold to select high-confidence predictions for updating the model.}

\B{The network processing process of multiview training strategy unfolds as follows: A remote sensing image $\x$ from the original dataset $\mathcal{D}$ is fed into the model. The corresponding prediction $\y$ is compared with the ground-truth label $\bm{T}$ to compute  the cross-entropy loss, serving as supervision for the first view. Simultaneously, $\y$ is threshold-filtered to generate a confidence map, which guides the supervision of the second view. Each pixel in the confidence map has a value of one if the corresponding pixel value exceeds a predefined threshold  $\tau$, otherwise, it is zero. Concurrently, $\x^{\text{aug}}$ is generated as the strongly augmented data, which is inputted into the model alongside $\x$. It results in corresponding resulting prediction $\y^{\text{aug}}$. The cross-entropy loss is calculated between $\y^{\text{aug}}$ and the ground-truth to obtain preliminary loss $\y^{\text{aug}}$, which is then multiplied element-wise by the confidence map. Finally, the weighted sum of two losses obtained from the multiview supervision is aggregated to form the overall loss used for training. This overall loss is then propagated backward through the network for parameter updates. The procedure is visually depicted in Fig.~\ref{MST}.}

The first one employs normal supervised learning loss function. Here, the normal image $\x$ is fed into HR-cloud-Net, and its resulting prediction $\y$ is compared with the ground-truth label $\bm{T}$ to compute the cross-entropy loss $\mathcal{L}_{\rm ce}$, defined as:
\begin{align}
	\mathcal L_{\rm ce}=-\sum_{i\in\mathcal P}\sum_{j\in \mathcal C} t_{ij}\log y_{ij}\label{loss_sup}
\end{align}
where $\y=[y_{ij}]$, $\bm T=[t_{ij}]$, $\mathcal C=\{1,2\},$ and $ \mathcal P=\{1,\ldots,W\times H\}$.

For the second loss function, we employ a multiview supervision strategy. In one perspective, termed the teacher view, we utilize the normal image as input, while in the other perspective, termed the student view, we utilize the augmented image. $\x^{\text{aug}}$ denotes the augmented image of $\x$, which is simultaneously fed into HR-cloud-Net alongside $\x$, resulting in corresponding outputs $\y^{\text{aug}}$. In this multiview supervision strategy, we process $\y$ through threshold filtering to generate a high-confidence mask, which guides the supervision of the student view. In the high-confidence mask, each pixel takes a value of one if the corresponding pixel value of $\y$ exceeds the threshold $\tau$; otherwise, it is set to zero. Subsequently, we compute the cross-entropy loss between $\y^{\text{aug}}$ and the ground-truth label $\bm{T}$ to define the second loss function, represented as:
\begin{align}
	\mathcal L_{\rm ce}^{\rm aug}=-\sum_{i\in\mathcal P}\sum_{j\in \mathcal C}
	\mathbb I(y_{ij}>\tau)
	t_{ij}\log y_{ij}^{\rm aug}\label{loss_w2s}
\end{align}
where $\mathbb I(\cdot)$ represents an indicator function that outputs one if the condition is true and zero otherwise. $\mathbb I(y_{ij}>\tau)$ returns the high-confidence mask. The mask allows for the selection of only the indices corresponding to high-confidence predictions in the teacher view, which are then utilized in updating the model parameters through the loss function $\mathcal{L}_{\text{ce}}^{\text{aug}}$. This mechanism ensures that the prediction of the noisy augmented inputs of the student view are guided by cleaner supervisions derived from the teacher view, particularly focusing on those with higher confidence. We regard the indices of these high-confidence predictions as valuable knowledge, facilitating its transfer to the student view during training.

The total loss for updating HR-cloud-Net's parameters is computed as the weighted sum of all losses obtained from the multiview supervision. This training procedure is visually depicted in Fig.~\ref{MST}. The overall loss function is defined by,
\begin{align}
	\mathcal{L}=\lambda _1\mathcal{L}_{\rm ce}+\lambda _2\mathcal{L}_{\rm ce}^{\rm aug}\label{loss_total}
\end{align}
where $\lambda_1$ and $\lambda_2$ are trade-off hyper parameters.
\section{Experimental Results}
\subsection{Experimental Setup}
We utilize images captured by the Landsat-8 satellite, sourced from the CHLandSat-8 dataset for training~\cite{du2023gated}. The CHLandSat-8 dataset comprises 64 panoramic scenes captured between January 2021 and December 2021, covering various regions in China. This dataset includes diverse land cover types such as urban areas, ice, snow, grasslands, mountains, forests, oceans, and deserts. The image size is $8000\times8000\times3$. We follow the approach proposed by Du \textit{et al.}~\cite{du2023gated}, utilizing 44 scenes from CHLandSat-8 as the training set.

Subsequently, to test the generalization capability of HR-cloud-Net, we evaluate HR-cloud-Net on three LandSat-8 datasets: CHlandSat-8~\cite{du2023gated}, 38-cloud~\cite{8898776}, and SPARCS~\cite{rs6064907}. The ground-truth label of 38-cloud mainly relies on function mask algorithms~\cite{zhu2012object,zhu2015improvement}, while SPARS is a high-quality dataset annotated manually~\cite{rs6064907}. The test set of CHLandSat-8 has the rest 20 scenes. 38-cloud has 20 scenes in its test set. The image size of CHLandSat-8 and 38-cloud is $8000\times 8000 \times3$. Similarly, we test on SPARCS~\cite{rs6064907}. SPARCS comprises patches of size $1000\times1000\times3$ extracted from 80 scenes. Detailed information about these three LandSat-8 test sets is shown in Table~\ref{tab:test_set}.
\B{
	CHLandsat-8 encompasses a wide range of complex scenarios, including urban areas, snow and ice, grasslands, mountains, forests, oceans, and deserts. 38-cloud includes diverse scenarios such as vegetation, bare soil, buildings, urban areas, water, snow, ice, and haze. SPARCS contains 80 pre-collected Landsat-8 scenario subsets with globally distributed sampled scenarios.  HR-cloud-Net is tested across these three datasets acquired from Landsat-8, demonstrating its strong generalization capabilities and its effectiveness in handling cloud detection tasks under various scenarios and conditions.
}
\begin{table}[!tp]
	\centering
	\begin{tabular}{lcc}
		\toprule
		The test set & Images  &  Scenes  \\
		\midrule
		CHLandSat-8 & 10080 & 20  \\
		38-cloud & 10906 & 20  \\
		SPARCS & 720 & 80  \\
		\bottomrule
	\end{tabular}
	\caption{The details of the three test sets.}
	\label{tab:test_set}
\end{table}

We uniformly crop the large images of the three datasets into small images of $352 \times 352$ for training, and use the same cropping method for testing to ensure experiment consistency. During training, in order to achieve the detection effect of real images and ensure integrity, we re-splice the cropped small images into normal large images for detection, facilitating comparison of the results and observation of image quality. We use random initialization to generate the weights of the backbone network, residual convolutional layers, and modules during network training. Optimization is performed using the ADAM optimizer~\cite{kingma2014adam} with an initial learning rate of 5e-5 and a weight decay rate of 0.0005. Training is set to run for 60 epochs with a batch size of eight. The threshold $\tau$ is set to 0.8, and the parameters $\lambda_1$ and $\lambda_2$ are set to 0.1.  In data augmentation, we focus on crafting a suite of data augmentation techniques finely tuned for images capturing cloud formations within the atmosphere. Leveraging a probabilistic framework, we meticulously calibrate alterations in brightness, contrast, saturation, and hue with a high likelihood of 80\%. Furthermore, we introduce randomness into the augmentation process by occasionally converting images to grayscale, with a probability of 20\%, and applying blur effects with a 50\% likelihood~\cite{sun2023corrmatch}. Through the strategic application of these augmentation strategies, we significantly bolster the HR-cloud-Net's capacity for generalization, paving the way for enhanced performance across diverse datasets.

The cloud detection performance is evaluated using three widely-used metrics: mean absolute error\cite{abs-2008-12134} ($e_{\rm ma}$), weighted $F_{\beta}$-measure\cite{6909433}
$(F_{\beta}^w)$, and structure measure\cite{abs-1708-00786} ($m_{\rm s}$).

The metric $e_{\rm ma}$ is used to measure the difference between the predicted mask and the ground-truth mask\cite{abs-2008-12134}. It evaluates the mean of the pixel-wise absolute error between the predicted cloud area and the ground-truth cloud mask, as defined by the equation:
\begin{equation}
	e_{\rm ma} = \frac1{|\mathcal P|} \sum_{i\in\mathcal P} |y_{i} - t_{i}|\,.
	\label{eq:mae}
\end{equation}
where ${|\mathcal P|}$ is the total number of pixels.

The $F_{\beta}^w$ metric represents the weighted ${F_\beta}$ measure, providing a comprehensive evaluation by considering both Precision and Recall\cite{6909433}. $F_{\beta}^w$ is computed as:
\begin{equation}
	F_{\beta}^w = \frac{{(1 + \beta^2)  {\rm Precision}^w  {\rm Recall}^w}}{{\beta^2 {\rm Precision}^w + {\rm Recall}^w}}
	\label{eq:fbeta}
\end{equation}
where ${\beta}$ serves as a trade-off parameter to tune the relative importance between Precision and Recall, and $w$ is the weighted parameter. Typically, when ${\beta > 1}$, Recall receives more emphasis, whereas for ${\beta < 1}$, Precision is more emphasis. When ${\beta = 1}$, both Precision and Recall are equally balanced. Precision assesses the accuracy of positive predicted pixels, calculated as the ratio of true positive predictions (correctly identified clouds) to all positive predictions (including true positives and false positives). Recall, also known as sensitivity or true positive rate, evaluates the completeness of positive predictions, defined as the ratio of true positive predictions to all actual positive pixels in the image.

The structure measure, $m_s$, is defined to evaluate assesses region-aware and object-aware structural similarity\cite{abs-1708-00786}. $m_{\rm s}$ is expressed as the weighted sum of ${S_o}$ and ${S_r}$, defined by,
\begin{equation}
	m_{\rm s} = \alpha S_o+ (1 - \alpha) S_r
	\label{eq:smeasure}
\end{equation}
where ${\alpha}$ serves as a trade-off parameter to adjust the relative weights of ${S_o}$ and ${S_r}$. ${S_o}$ represents the perceptual structural similarity of the object, computed as the weighted average of the structural similarities~\cite{1284395} of the foreground and background images. ${S_r}$ represents the adjustment of perceptual similarity in regions, obtained by dividing the image into several regions and computing the structural similarity~\cite{1284395} of each region.
\subsection{Comparative Experiments}
We conduct a comprehensive comparison of HR-cloud-Net with several existing network models, including U-Net~\cite{ronneberger2015u}, PSPNet~\cite{zhao2017pyramid}, SegNet\cite{7803544}, Cloud-Net\cite{8898776}, CDNet~\cite{yang2019cdnet}, CDNet-v2~\cite{9094671}, HRNet~\cite{hrnet8953615,wang2020deep}, and GANet~\cite{du2023gated}. To ensure consistency, all models are trained on the CHLandSat-8 train set and evaluated on the CHLandSat-8 test set, 38-cloud test set, and SPARCS dataset using the same setting.
\subsubsection{Quantitative Experiments}
We conduct comparative experiments to assess the quality of segmented cloud areas. Multiple evaluation metrics, including $e_{\rm ma}$, $F_{\beta}^w$, and $m_s$, are employed to evaluate the performance of the models. In Tables~\ref{tab:chlandsat}, \ref{tab:38-cloud}, and \ref{tab:SPARCS}, the best-performing metrics are highlighted in bold, while the second best are in italics. This allows for easy comparison of performance between different networks. The quantitative results demonstrate that HR-cloud-Net achieves comparable performance to state-of-the-art baselines across all three evaluation metrics and datasets.

Table~\ref{tab:chlandsat} presents the evaluation results from the CHLandSat-8 test set. HR-cloud-Net demonstrates substantial advantages across all evaluation metrics, showcasing its ability to accurately extract cloud features and generate precise cloud region masks. Similarly, in Table~\ref{tab:38-cloud}, HR-cloud-Net achieves the best performance in the 38-cloud test dataset. In Table~\ref{tab:SPARCS}, HR-cloud-Net exhibits significant advantages across all evaluation metrics in the SPARCS dataset.

Tables~\ref{tab:chlandsat}, \ref{tab:38-cloud}, and \ref{tab:SPARCS} clearly demonstrates that HR-cloud-Net exhibits superior capability in feature extraction and yields higher-quality cloud mask segmentation images. It is worth noting that HR-cloud-Net exhibits outstanding performance across all three datasets in the $F_{\beta}^w$ metric. The $F_{\beta}^w$ metric places significant emphasis on cloud edge conditions, assigning them higher metric.

\begin{table}[t]
	\centering
	\begin{tabular}{l|ccc}
		\toprule
		Method & $e_{\rm ma}$$\downarrow$ & ${F_{\beta}^w}$ $\uparrow$ & $m_s\uparrow$ \\
		\midrule
		U-Net & 0.1130 & 0.7448  & 0.7288   \\
		PSPNet & 0.0969  & 0.7989  & 0.7672  \\
		SegNet & 0.1023  & 0.7780 & 0.7540  \\
		Cloud-Net & 0.1012  & 0.7641  & 0.7368   \\
		CDNet& 0.1286  & 0.7222  &  0.7087 \\
		CDNet-v2&  0.1254 & 0.7350  &  0.7141 \\
		HRNet& \emph{\textbf{0.0737}}&  0.8297  &  \emph{\textbf{0.8184}}  \\
		GANet& 0.0751  & \emph{\textbf{0.8396}}  &  0.8106  \\
		HR-cloud-Net&   \textbf{0.0628} &  \textbf{0.8503}  &  \textbf{0.8337}   \\
		\bottomrule
	\end{tabular}
	\caption{Comparison of HR-cloud-Net with baselines on CHLandSat-8 dataset.}
	\label{tab:chlandsat}
\end{table}

\begin{table}[t]
	\centering
	\begin{tabular}{l|ccc}
		\toprule
		Method & $e_{\rm ma}$$\downarrow$  &  ${F_{\beta}^w}$ $\uparrow$  &  $m_s\uparrow$ \\
		\midrule
		U-Net & 0.0638  &  0.7966 & 0.7845  \\
		PSPNet & 0.0653  & 0.7592  &  0.7766 \\
		SegNet & 0.0556  & 0.8002  & 0.8059 \\
		Cloud-Net & 0.0556  &  0.7615 & 0.7987 \\
		CDNet&  0.1057 & 0.7378  &  0.7270 \\
		CDNet-v2& 0.1084  &  0.7183 &  0.7213\\
		HRNet& 0.0538 &  0.8086 & 0.8183 \\
		GANet& \emph{\textbf{0.0410}} & \emph{\textbf{0.8159}}  & \emph{\textbf{0.8342}} \\
		HR-cloud-Net& {\textbf{0.0395}} &  \textbf{0.8673 }  &  \textbf{ 0.8479}  \\
		\bottomrule
	\end{tabular}
	\caption{Comparison of HR-cloud-Net with baselines on on 38-cloud dataset.}
	\label{tab:38-cloud}
\end{table}

\begin{table}[t]
	\centering
	\begin{tabular}{l|ccc}
		\toprule
		Method & $e_{\rm ma}$$\downarrow$  &  ${F_{\beta}^w}$ $\uparrow$  &  $m_s\uparrow$ \\
		\midrule
		U-Net & 0.1314  & 0.3651  & 0.5416  \\
		PSPNet & 0.1263  & 0.3758  & 0.5414 \\
		SegNet & 0.1100  & 0.4697  & 0.5918 \\
		Cloud-Net & 0.1213  & 0.3804  & 0.5536 \\
		CDNet&  0.1157 &  0.4585 & 0.5919 \\
		CDNet-v2& 0.1219  &  0.4247 &  0.5704\\
		HRNet& 0.1008  &  0.3742 &  0.5777\\
		GANet&\emph{\textbf{0.0987}} &\emph{\textbf{0.5134}}&\emph{\textbf{0.6210}}\\
		HR-cloud-Net&  \textbf{0.0833}  &  \textbf{0.5202} &  \textbf{0.6327} \\
		\bottomrule
	\end{tabular}
	\caption{Comparison of HR-cloud-Net with baselines on SPARCS dataset.}
	\label{tab:SPARCS}
\end{table}
\subsubsection{Qualitative Experiments}
This paper presents qualitative comparison results between HR-cloud-Net and several comparative algorithms, including U-Net~\cite{ronneberger2015u}, PSPNet~\cite{zhao2017pyramid}, SegNet\cite{7803544}, Cloud-Net\cite{8898776}, CDNet~\cite{yang2019cdnet}, CDNet-v2~\cite{9094671}, HRNet~\cite{hrnet8953615,wang2020deep}, and GANet~\cite{du2023gated}. These comparisons aim to validate the superior performance of the proposed method through more intuitive assessments. The qualitative results underscore HR-cloud-Net's ability to more accurately capture cloud texture information and generate precise cloud detection results.

\begin{figure*}[!t]
	\centering
	\includegraphics[width=17cm,height=19cm]{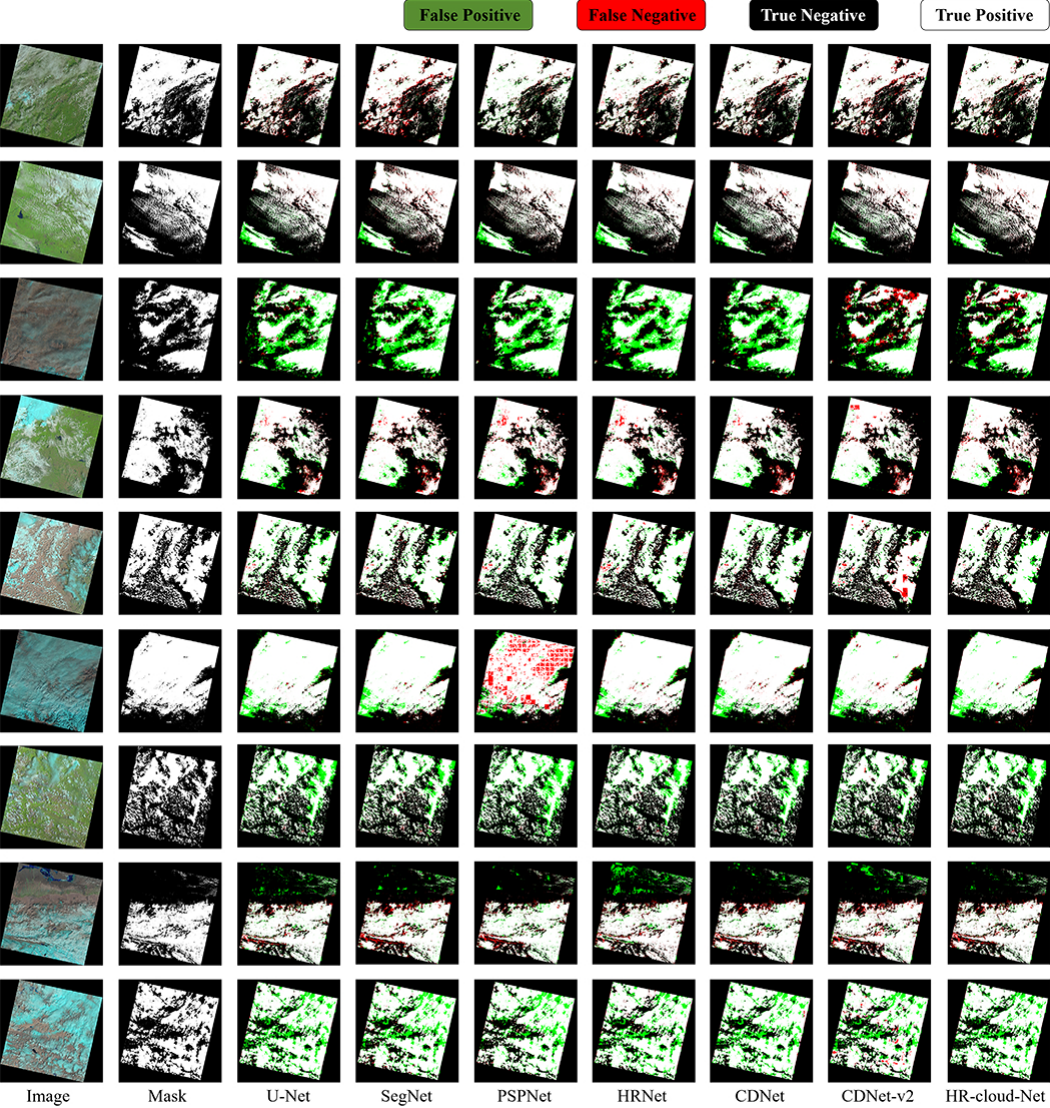}
	\caption{Visual comparison of HR-cloud-Net with other methods on the CHLandSat-8 dataset.}
	\label{CHLandsat}
\end{figure*}

\begin{figure*}[!t]
	\centering
	\includegraphics[width=17cm,height=19cm]{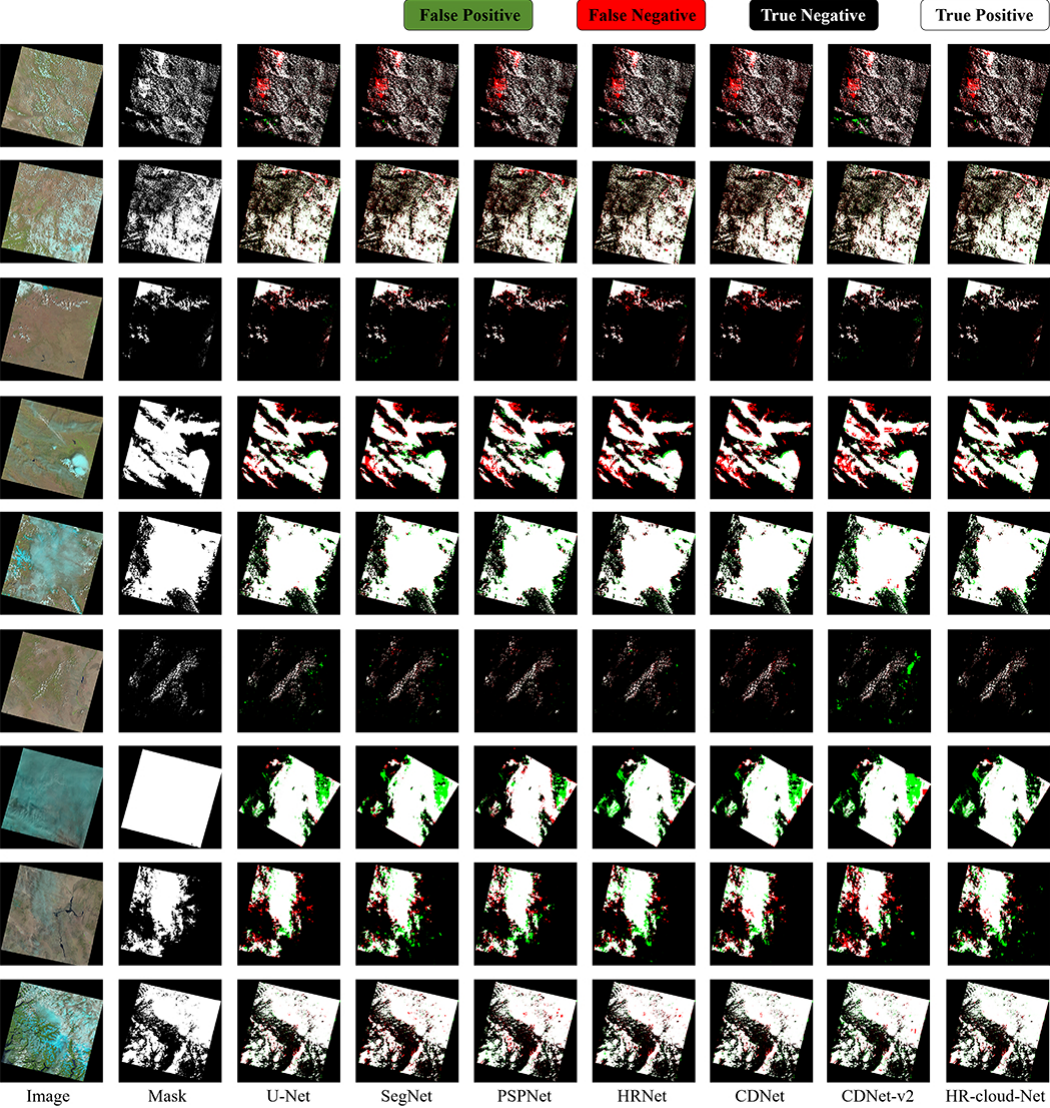}
	\caption{Visual comparison of HR-cloud-Net with other methods on the 38-cloud dataset.}
	\label{38-cloud}
\end{figure*}
To facilitate a more intuitive observation of the cloud detection performance across different networks, we generate cloud mask images produced by each network and compare them with the original dataset images alongside their respective cloud mask images. The images generated by different networks are arranged chronologically based on their release time.

\begin{figure*}[!t]
	\centering
	\includegraphics[width=17cm,height=19cm]{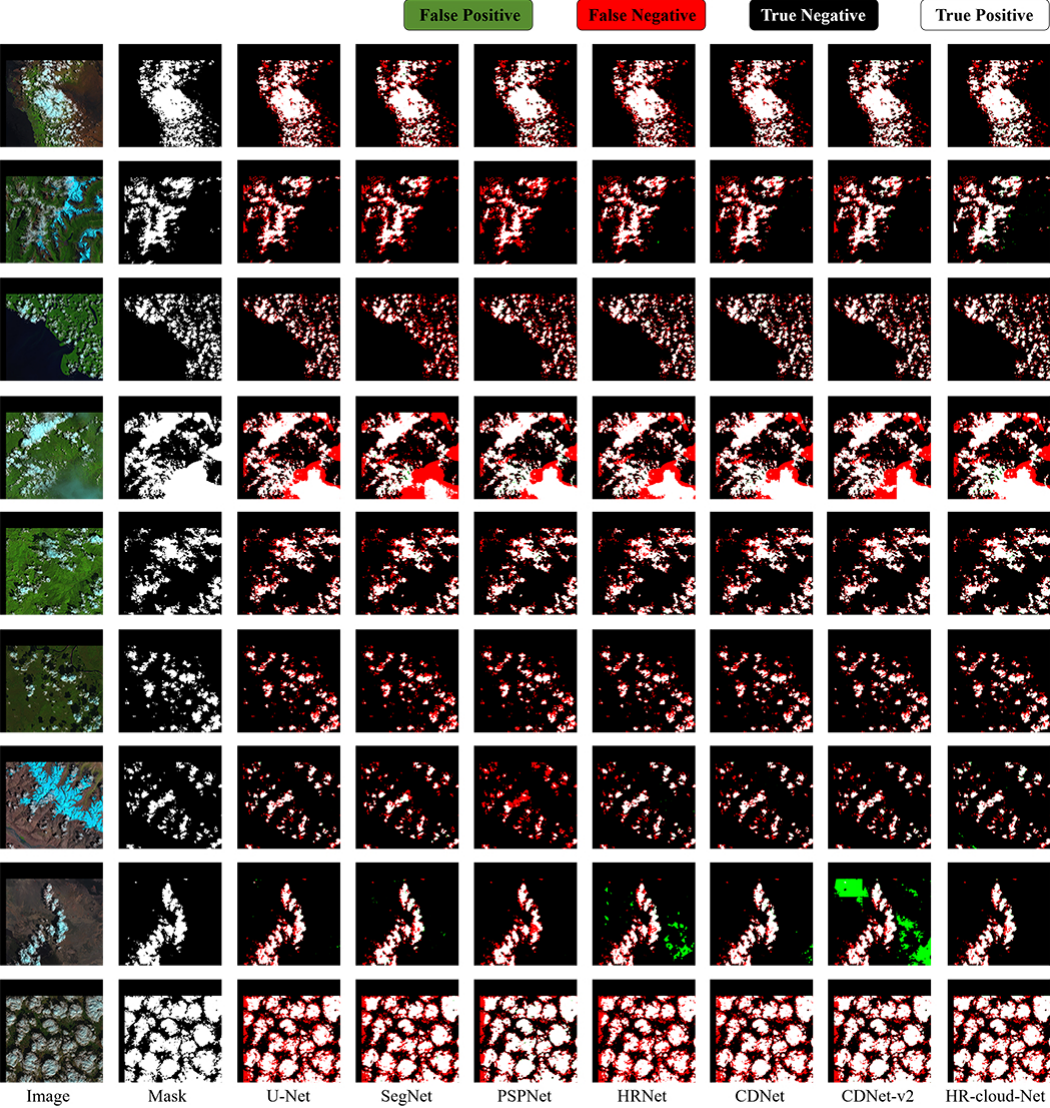}
	\caption{Visual comparison of HR-cloud-Net with other methods on the SPARCS dataset.}
	\label{SPARCS}
\end{figure*}

As shown in Figs.~\ref{CHLandsat}, \ref{38-cloud}, and \ref{SPARCS}, HR-cloud-Net effectively manages various types of cloud areas, yielding precise cloud detection results. In the subsequent comparison of Figs.~\ref{CHLandsat}, \ref{38-cloud}, and \ref{SPARCS}, False Positive denotes a non-cloud region incorrectly detected as a cloud region. Conversely, False Negative denotes a cloud region erroneously identified as a non-cloud region. True Negative indicates a non-cloud region correctly detected as a non-cloud region, while True Positive represents a cloud region correctly detected as a cloud region.

The HR-cloud-Net effectively handles remote sensing images from various scenes by better learning the texture features of cloud layers. We carefully select different real-world scenes from three datasets: CHLandSat-8, 38-cloud, and SPARS, each containing nine images. As shown in Fig.~\ref{CHLandsat}, the scenes in the CHLandSat-8 test dataset are primarily distributed in plain areas. Previous algorithms often overlooked high-resolution texture features, resulting in significant issues of missed detection and false alarms when dealing with complex remote sensing images from different scenes. However, HR-cloud-Net is capable of leveraging richer texture features, leading to excellent performance across diverse scenes. Fig.~\ref{38-cloud} illustrates the cloud detection results from desert, plateau, and snow scenes selected from the 38-cloud test set. For the SPARCS dataset, we chose ocean, snow, and cloud shadow scenes for evaluation. As shown in Fig.~\ref{SPARCS}, HR-cloud-Net achieves superior visual detection results in these scenes as well.

\begin{figure*}[!t]
	\centering
	\includegraphics[width=0.94\textwidth]
	{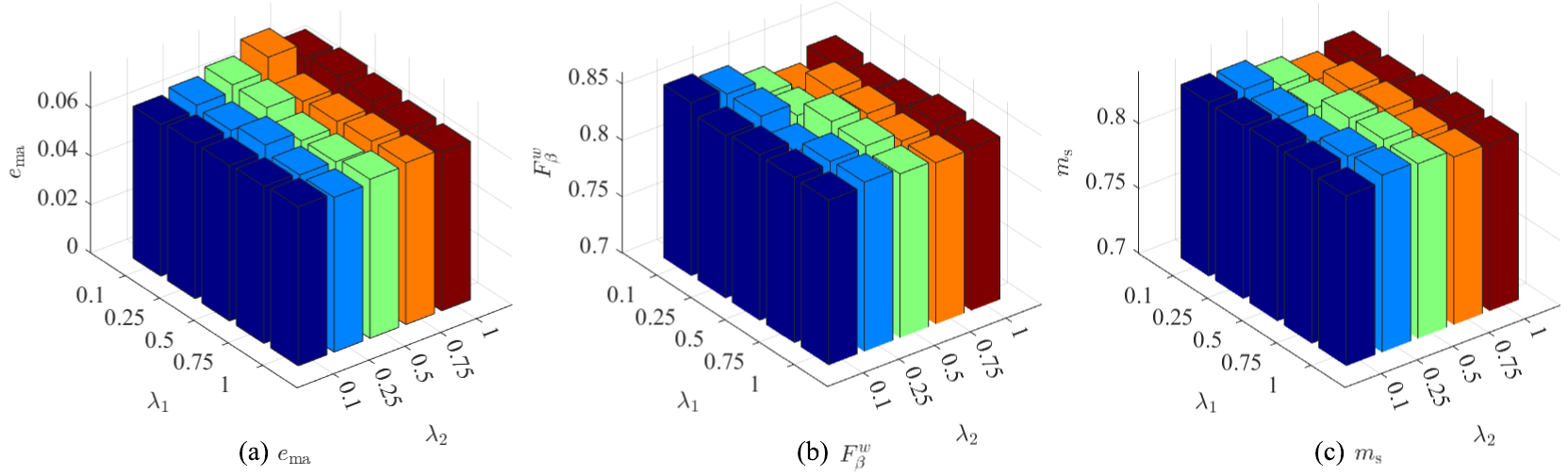}
	\caption{\B{Sensitivity analysis on $\lambda_1$ and $\lambda_2$.}} \label{fig:para}
\end{figure*}
The subjective comparison results of different object detection algorithms indicate that the high-resolutio representation module designed in HR-cloud-Net effectively integrates high-resolution texture features and low-resolution positional features, thereby alleviating the problem of unclear and inaccurate texture features in images and assisting in predicting clear image outputs. Additionally, the layer-wise cascaded feature fusion module and the multi-resolution pyramid pooling module adjust features of different resolutions, significantly enhancing the aggregation effect across different resolutions. Through multiview supervision, HR-cloud-Net better adapts to images captured under various conditions. Subjective results suggest that HR-cloud-Net partially mitigates the issue of inaccurate edge prediction encountered by existing  detection algorithms.

In subjective comparison results, HR-cloud-Net demonstrates fewer false negative cases across all datasets, meaning fewer instances where cloud regions are incorrectly classified as non-cloud areas, and vice versa. This discrepancy is attributed to the expression of texture features. Lower-resolution structures require more interpolation to restore the original image size, potentially leading to areas that should be clouds being mistakenly labeled as non-cloud areas. In contrast, HR-cloud-Net places greater emphasis on high-resolution feature extraction, enabling better restoration of cloud texture features and thereby reducing the misclassification of cloud regions as non-cloud areas. Particularly noteworthy is the superior performance of HR-cloud-Net in texture handling, especially evident in the more challenging SPARCS test dataset.

\B{As defined by Eq.~\eqref{loss_total}, the loss function of HR-cloud-Net consists of two terms, each with its own trade-off parameter. For HR-cloud-Net, parameter insensitivity is crucial for enhancing stability. The two parameters, $\lambda_1$ and $\lambda_2$, need to be determined. To illustrate the insensitivity to variations in $\lambda_1$ and $\lambda_2$, we conducted experiments measuring $e_{\rm ma}$, $F_{\beta}^w$, and $m_{\rm s}$ as shown in Figure~\ref{fig:para}. Both parameters were varied within the range of [0.1, 0.25, 0.5, 0.75, 1]. The results indicate stable performance across a wide range of parameter values, with the optimal results achieved when both $\lambda_1$ and $\lambda_2$ are set to 0.1.}

HR-cloud-Net achieves the comparable performance through intuitive image comparison. Mainly because other methods focus more on regional accuracy, thereby neglecting the processing of cloud edges, and the low-resolution information extraction they adopt may cause useful information to be overwhelmed, resulting in many false detection. HR-cloud-Net we designed not only ensures the correct transmission of semantic information between low-resolution and high-resolution features, but also applies the multi-resolution pyramid pooling module to enhance HR-cloud-Net's ability to capture global information. This gives HR-cloud-Net better performance in areas prone to error detection, better ability to handle cloud edges, and more accurate cloud detection.
\subsection{Ablation Study}
To ascertain the accuracy and efficiency of HR-cloud-Net, we divide the ablation experiment into two parts. The first part assesses the influence of various modules within HR-cloud-Net on the overall network. The second part examines the effects of incorporating different loss functions into HR-cloud-Net. We conduct cloud prediction evaluations on the CHLandSat-8 dataset using $e_{\rm ma}$, ${F_{\beta}^w}$, and $m_s$ metrics, ensuring consistency across all other algorithm configurations.

To assess the precision and efficiency of different modules of HR-cloud-Net, we conduct a series of ablation experiments, sequentially disabling different modules. The experimental procedure proceeded as follows: First, we evaluate performance of HR-cloud-Net without the cascaded feature fusion. Second, we conduct experiments without the multi-resolution pyramid pooling module. Third, we remove multiple resolution, i.e., we retain the highest resolution from stage 4  while we removing some ineffective skip connections in stage 4 and cascaded feature fusion module. The results, presented in Table~\ref{tab:example}, demonstrate that the high-resolution branch we designed has effectively learned the characteristics of clouds, but multi-resolution, as a useful complement, improves the accuracy of cloud detection. Meanwhile, both the cascaded feature fusion and the multi-resolution pyramid pooling module contribute to enhancing the model's expressive capacity. Furthermore, our layer-wise cascaded feature fusion and multi-resolution pyramid pooling module effectively improve HR-cloud-Net performance. 

\begin{table}[!t]
	\centering
	\begin{tabular}{l|ccc}
		\toprule
		& $e_{\rm ma}\,\downarrow$  & $F_{\beta}^w\,\uparrow$  &  $m_s\,\uparrow$ \\
		\midrule
		w/o Cascaded Feature Fusion &0.0754  &0.8292  & 0.8104 \\
		w/o Pyramid Pooling Module&0.0655  &0.8463 &0.8313  \\
		w/o  Multiple Resolutions &0.0658  &0.8429   &0.8292\\
		HR-cloud-Net& \textbf{0.0628} &  \textbf{0.8503}  &  \textbf{0.8337}  \\
		\bottomrule
	\end{tabular}
	\caption{Ablation study of different modules.}
	\label{tab:example}
\end{table}

\begin{table}[!t]
	\centering
	\begin{tabular}{cc|ccc}
		\toprule
		${\mathcal L_{\rm ce}}$ & $\mathcal L_{\rm ce}^{\rm aug}$ & $e_{\rm ma}$$\downarrow$  & ${F_{\beta}^w}$ $\uparrow$  &  $m_s$$\uparrow$ \\
		\midrule
		\checkmark &   & 0.0711 & 0.8344 &  0.8209 \\
		\checkmark &\checkmark&  \textbf{0.0628} &  \textbf{0.8503}  &  \textbf{0.8337}  \\
		\bottomrule
	\end{tabular}
	\caption{Ablation of different loss functions.}
	\label{Ablation}
\end{table}

To assess the efficacy of our multiview supervision approach, we conducted ablation experiments, scrutinizing both the conventional supervised loss and our novel multiview supervision method. Specifically, we investigated the impact of solely employing the supervised loss compared to its integration with our multiview supervision training strategy. These experiments were conducted on the CHLandSat-8 dataset, employing established evaluation metrics such as $e_{\rm ma}$, $F_{\beta}^w$, and $m_s$. The findings, presented in Table~\ref{Ablation}, highlight the effectiveness of our optimized multiview supervision training strategy, showcasing substantial enhancements in algorithmic performance.
\section{Conclusion}
HR-cloud-Net is designed using a hierarchical high-resolution integration approach for cloud detection in remote sensing images. This architecture ingeniously integrates features of high-resolution representation, parallel multi-resolution flows, and pyramid pooling for semantic information aggregation. We introduce a sophisticated data augmentation strategy, where the student view, trained on meticulously augmented images, undergoes supervision by the teacher view, trained on pristine normal images. This approach provides a robust and adaptive solution tailored for cloud detection tasks, accentuating the considerable potential for attaining proficient generalization, particularly in settings where data availability is limited. This ensures versatility and reliability in various operational environments. Experimental results demonstrate that HR-cloud-Net achieves optimal performance across various evaluation metrics. This further validates the outstanding performance of HR-cloud-Net in cloud detection tasks, offering a straightforward yet effective solution for future advancements in remote sensing image analysis and related domains.
\section*{Code, Data, and Materials Availability}
The source code is available at \url{https://github.com/kunzhan/HR-cloud-Net}. The download links for the three datasets are provided in README.md of the GitHub repository.
\section*{Acknowledgments}
This work was supported by the Fundamental Research Funds for the Central Universities under No.~lzujbky-2022-ct06 and Supercomputing Center of Lanzhou University. We thank Xianjun Du for the help of discussion.
{
    \small
    \bibliographystyle{ieeenat_fullname}
    \bibliography{main}
}


\end{document}

%% file: preamble.tex
\usepackage[dvipsnames]{xcolor}


%% file: main.bbl
\begin{thebibliography}{32}
\providecommand{\natexlab}[1]{#1}
\providecommand{\url}[1]{\texttt{#1}}
\expandafter\ifx\csname urlstyle\endcsname\relax
  \providecommand{\doi}[1]{doi: #1}\else
  \providecommand{\doi}{doi: \begingroup \urlstyle{rm}\Url}\fi

\bibitem[Badrinarayanan et~al.(2017)Badrinarayanan, Kendall, and
  Cipolla]{7803544}
Vijay Badrinarayanan, Alex Kendall, and Roberto Cipolla.
\newblock {SegNet}: A deep convolutional encoder-decoder architecture for image
  segmentation.
\newblock \emph{IEEE Transactions on Pattern Analysis and Machine
  Intelligence}, 39\penalty0 (12):\penalty0 2481--2495, 2017.

\bibitem[Chen et~al.(2023)Chen, Xia, Lin, and Qian]{10144804}
Kai Chen, Min Xia, Haifeng Lin, and Ming Qian.
\newblock Multiscale attention feature aggregation network for cloud and cloud
  shadow segmentation.
\newblock \emph{IEEE Transactions on Geoscience and Remote Sensing},
  61:\penalty0 1--16, 2023.

\bibitem[Du and Wu(2024)]{du2023gated}
Xianjun Du and Hailei Wu.
\newblock Gated aggregation network for cloud detection in remote sensing
  image.
\newblock \emph{The Visual Computer}, 40:\penalty0 2517--2536, 2024.

\bibitem[Fan et~al.(2017)Fan, Cheng, Liu, Li, and Borji]{abs-1708-00786}
Deng{-}Ping Fan, Ming{-}Ming Cheng, Yun Liu, Tao Li, and Ali Borji.
\newblock Structure-measure: {A} new way to evaluate foreground maps.
\newblock In \emph{ICCV}, pages 4548--4557, 2017.

\bibitem[Fu et~al.(2022)Fu, Fan, Ji, Zhao, Shen, and Zhu]{abs-2008-12134}
K. Fu, D. Fan, G. Ji, Q. Zhao, J. Shen, and C. Zhu.
\newblock Siamese network for {RGB-D} salient object detection and beyond.
\newblock \emph{IEEE Transactions on Pattern Analysis and Machine
  Intelligence}, 44\penalty0 (09):\penalty0 5541--5559, 2022.

\bibitem[Guo et~al.(2021)Guo, Yang, Yue, Tan, Hou, and Li]{9094671}
Jianhua Guo, Jingyu Yang, Huanjing Yue, Hai Tan, Chunping Hou, and Kun Li.
\newblock {CDNet-v2}: Cnn-based cloud detection for remote sensing imagery with
  cloud-snow coexistence.
\newblock \emph{IEEE Transactions on Geoscience and Remote Sensing},
  59\penalty0 (1):\penalty0 700--713, 2021.

\bibitem[He et~al.(2016)He, Zhang, Ren, and Sun]{he2016deep}
Kaiming He, Xiangyu Zhang, Shaoqing Ren, and Jian Sun.
\newblock Deep residual learning for image recognition.
\newblock In \emph{CVPR}, pages 770--778, 2016.

\bibitem[Hu et~al.(2021)Hu, Zhang, and Xia]{rs13224533}
Kai Hu, Dongsheng Zhang, and Min Xia.
\newblock {CDU-Net}: Cloud detection {U-Net} for remote sensing imagery.
\newblock \emph{Remote Sensing}, 13\penalty0 (22):\penalty0 4533, 2021.

\bibitem[Hu et~al.(2023)Hu, Zhang, Xia, Weng, and Lin]{hu2023mcanet}
Kai Hu, Enwei Zhang, Min Xia, Liguo Weng, and Haifeng Lin.
\newblock {MCANet}: A multi-branch network for cloud/snow segmentation in
  high-resolution remote sensing images.
\newblock \emph{Remote Sensing}, 15\penalty0 (4):\penalty0 1055, 2023.

\bibitem[Huang et~al.(2020)Huang, Lin, Tong, Hu, Zhang, Iwamoto, Han, Chen, and
  Wu]{huang2020Unet}
Huimin Huang, Lanfen Lin, Ruofeng Tong, Hongjie Hu, Qiaowei Zhang, Yutaro
  Iwamoto, Xianhua Han, Yen-Wei Chen, and Jian Wu.
\newblock {U-Net3+}: A full-scale connected {U-Net} for medical image
  segmentation.
\newblock In \emph{ICASSP}, pages 1055--1059, 2020.

\bibitem[Hughes and Hayes(2014)]{rs6064907}
M.~Joseph Hughes and Daniel~J. Hayes.
\newblock Automated detection of cloud and cloud shadow in single-date landsat
  imagery using neural networks and spatial post-processing.
\newblock \emph{Remote Sensing}, 6\penalty0 (6):\penalty0 4907--4926, 2014.

\bibitem[Jeppesen et~al.(2019)Jeppesen, Jacobsen, Inceoglu, and
  Toftegaard]{jeppesen2019cloud}
Jacob~H{\o}xbroe Jeppesen, Rune~Hylsberg Jacobsen, Fadil Inceoglu, and
  Thomas~Skj{\o}deberg Toftegaard.
\newblock A cloud detection algorithm for satellite imagery based on deep
  learning.
\newblock \emph{Remote Sensing of Environment}, 229:\penalty0 247--259, 2019.

\bibitem[King et~al.(2013)King, Platnick, Menzel, Ackerman, and
  Hubanks]{king2013spatial}
Michael~D King, Steven Platnick, W~Paul Menzel, Steven~A Ackerman, and Paul~A
  Hubanks.
\newblock Spatial and temporal distribution of clouds observed by modis onboard
  the terra and aqua satellites.
\newblock \emph{IEEE Transactions on Geoscience and Remote Sensing},
  51\penalty0 (7):\penalty0 3826--3852, 2013.

\bibitem[Kingma and Ba(2015)]{kingma2014adam}
Diederik~P Kingma and Jimmy Ba.
\newblock Adam: A method for stochastic optimization.
\newblock In \emph{ICLR}, 2015.

\bibitem[Li et~al.(2018)Li, Shen, Wei, Cheng, and Yuan]{li2018cloud}
Zhiwei Li, Huanfeng Shen, Yancong Wei, Qing Cheng, and Qiangqiang Yuan.
\newblock Cloud detection by fusing multi-scale convolutional features.
\newblock In \emph{ISPRS Annals of the Photogrammetry, Remote Sensing and
  Spatial Information Sciences}, pages 149--152, 2018.

\bibitem[Margolin et~al.(2014)Margolin, Zelnik-Manor, and Tal]{6909433}
Ran Margolin, Lihi Zelnik-Manor, and Ayellet Tal.
\newblock How to evaluate foreground maps.
\newblock In \emph{CVPR}, pages 248--255, 2014.

\bibitem[Mohajerani and Saeedi(2019)]{8898776}
Sorour Mohajerani and Parvaneh Saeedi.
\newblock {Cloud-Net}: An end-to-end cloud detection algorithm for landsat 8
  imagery.
\newblock In \emph{IGARSS}, pages 1029--1032, 2019.

\bibitem[Ronneberger et~al.(2015)Ronneberger, Fischer, and
  Brox]{ronneberger2015u}
Olaf Ronneberger, Philipp Fischer, and Thomas Brox.
\newblock {U-Net}: Convolutional networks for biomedical image segmentation.
\newblock In \emph{MICCAI}, pages 234--241, 2015.

\bibitem[Shi et~al.(2019)Shi, Qi, Liu, Niu, and Zhang]{rs11222719}
Yan Shi, Zhixin Qi, Xiaoping Liu, Ning Niu, and Hui Zhang.
\newblock Urban land use and land cover classification using multisource remote
  sensing images and social media data.
\newblock \emph{Remote Sensing}, 11\penalty0 (22):\penalty0 2719, 2019.

\bibitem[Sun et~al.(2024)Sun, Yang, Zhang, Cheng, and Hou]{sun2023corrmatch}
Boyuan Sun, Yuqi Yang, Le Zhang, Ming-Ming Cheng, and Qibin Hou.
\newblock {CorrMatch}: Label propagation via correlation matching for
  semi-supervised semantic segmentation.
\newblock In \emph{CVPR}, 2024.

\bibitem[Sun et~al.(2019)Sun, Xiao, Liu, and Wang]{hrnet8953615}
Ke Sun, Bin Xiao, Dong Liu, and Jingdong Wang.
\newblock Deep high-resolution representation learning for human pose
  estimation.
\newblock In \emph{CVPR}, pages 5686--5696, 2019.

\bibitem[Wang et~al.(2021)Wang, Sun, Cheng, Jiang, Deng, Zhao, Liu, Mu, Tan,
  Wang, Liu, and Xiao]{wang2020deep}
Jingdong Wang, Ke Sun, Tianheng Cheng, Borui Jiang, Chaorui Deng, Yang Zhao,
  Dong Liu, Yadong Mu, Mingkui Tan, Xinggang Wang, Wenyu Liu, and Bin Xiao.
\newblock Deep high-resolution representation learning for visual recognition.
\newblock \emph{IEEE Transactions on Pattern Analysis and Machine
  Intelligence}, 43\penalty0 (10):\penalty0 3349--3364, 2021.

\bibitem[Wang et~al.(2004)Wang, Bovik, Sheikh, and Simoncelli]{1284395}
Zhou Wang, A.C. Bovik, H.R. Sheikh, and E.P. Simoncelli.
\newblock Image quality assessment: from error visibility to structural
  similarity.
\newblock \emph{IEEE Transactions on Image Processing}, 13\penalty0
  (4):\penalty0 600--612, 2004.

\bibitem[Yang et~al.(2019)Yang, Guo, Yue, Liu, Hu, and Li]{yang2019cdnet}
Jingyu Yang, Jianhua Guo, Huanjing Yue, Zhiheng Liu, Haofeng Hu, and Kun Li.
\newblock {CDNet}: {CNN}-based cloud detection for remote sensing imagery.
\newblock \emph{IEEE Transactions on Geoscience and Remote Sensing},
  57\penalty0 (8):\penalty0 6195--6211, 2019.

\bibitem[Yin et~al.(2022)Yin, Wang, Ni, and Hao]{Yin2022}
Meijie Yin, Peng Wang, Cui Ni, and Weilong Hao.
\newblock Cloud and snow detection of remote sensing images based on improved
  {U-Net3+}.
\newblock \emph{Scientific Reports}, 12\penalty0 (1):\penalty0 14415, 2022.

\bibitem[Zhang et~al.(2023)Zhang, Weng, Ding, Xia, and Lin]{zhang2023crsnet}
Chao Zhang, Liguo Weng, Li Ding, Min Xia, and Haifeng Lin.
\newblock {CRSNet}: Cloud and cloud shadow refinement segmentation networks for
  remote sensing imagery.
\newblock \emph{Remote Sensing}, 15\penalty0 (6):\penalty0 1664, 2023.

\bibitem[Zhang et~al.(2004)Zhang, Rossow, Lacis, Oinas, and
  Mishchenko]{zhang2004calculation}
Yuanchong Zhang, William~B Rossow, Andrew~A Lacis, Valdar Oinas, and Michael~I
  Mishchenko.
\newblock Calculation of radiative fluxes from the surface to top of atmosphere
  based on {ISCCP} and other global data sets: Refinements of the radiative
  transfer model and the input data.
\newblock \emph{Journal of Geophysical Research: Atmospheres}, 109\penalty0
  (D19), 2004.

\bibitem[Zhao et~al.(2017)Zhao, Shi, Qi, Wang, and Jia]{zhao2017pyramid}
Hengshuang Zhao, Jianping Shi, Xiaojuan Qi, Xiaogang Wang, and Jiaya Jia.
\newblock Pyramid scene parsing network.
\newblock In \emph{CVPR}, pages 2881--2890, 2017.

\bibitem[Zhou et~al.(2018)Zhou, Rahman~Siddiquee, Tajbakhsh, and
  Liang]{zhou2018Unet++}
Zongwei Zhou, Md~Mahfuzur Rahman~Siddiquee, Nima Tajbakhsh, and Jianming Liang.
\newblock {U-Net++}: A nested {U-Net} architecture for medical image
  segmentation.
\newblock In \emph{MICCAI}, pages 3--11, 2018.

\bibitem[Zhu and Woodcock(2012)]{zhu2012object}
Zhe Zhu and Curtis~E Woodcock.
\newblock Object-based cloud and cloud shadow detection in landsat imagery.
\newblock \emph{Remote Sensing of Environment}, 118:\penalty0 83--94, 2012.

\bibitem[Zhu and Woodcock(2014)]{zhu2014continuous}
Zhe Zhu and Curtis~E Woodcock.
\newblock Continuous change detection and classification of land cover using
  all available landsat data.
\newblock \emph{Remote Sensing of Environment}, 144:\penalty0 152--171, 2014.

\bibitem[Zhu et~al.(2015)Zhu, Wang, and Woodcock]{zhu2015improvement}
Zhe Zhu, Shixiong Wang, and Curtis~E Woodcock.
\newblock Improvement and expansion of the fmask algorithm: Cloud, cloud
  shadow, and snow detection for {Landsats 4--7, 8, and Sentinel} 2 images.
\newblock \emph{Remote Sensing of Environment}, 159:\penalty0 269--277, 2015.

\end{thebibliography}
